%% file: main.tex
\documentclass{article}
\usepackage{arxiv}
\usepackage[T1]{fontenc}
\usepackage[utf8]{inputenc}
\usepackage{microtype}

\input{math_commands.tex}

\usepackage[authoryear,round]{natbib}
\setcitestyle{authoryear,round,citesep={;},aysep={,},yysep={;}}
\usepackage{url}
\usepackage{booktabs}
\usepackage{graphicx}
\usepackage{float}
\usepackage{subcaption}
\usepackage{wrapfig}
\usepackage{multirow}
\usepackage{enumitem}
\setlist[itemize]{labelindent=0.8em,leftmargin=*,nosep}
\usepackage[hidelinks]{hyperref}
\hypersetup{
  pdftitle={DUET: Dual-Sparse Inference for Block Diffusion Language Models},
  pdfauthor={Siyuan Yan, Shen Gu, Jingwei Xu}
}

\title{DUET: Dual-Sparse Inference for \\ Block Diffusion Language Models}
\renewcommand{\shorttitle}{DUET: Dual-Sparse Inference for Block DLMs}
\author{Siyuan Yan \quad Shen Gu \quad Xiaoxing Ma \quad Chun Cao \quad  Jingwei Xu\thanks{Corresponding author.} \\
State Key Laboratory for Novel Software Technology, Nanjing University, China \\
\texttt{jingweix@nju.edu.cn}}
\date{}

\begin{document}
\maketitle

\begin{abstract}
Diffusion language models (DLMs) enable parallel token generation through iterative denoising, yet their inference remains costly due to repeated attention over a growing historical prefix and repeated processing of the entire current generation block. Historical attention becomes increasingly expensive as generation progresses, while all current-block positions are propagated through the Transformer at every denoising step even though only a small fraction are typically resolved at that step. In this paper, we introduce Duet, a training-free dual-sparse inference framework that addresses these challenges through prefix Key-Value (KV) sparsity and early query sparsity, respectively. Specifically, Duet exploits the concentration and cross-step stability of historical attention within each generation block to construct a compact historical KV subset that is reused across denoising steps. For query sparsity, Duet predicts update positions from shallow representations and propagates only the corresponding queries through the suffix Transformer layers, avoiding deep computation at positions unlikely to be updated in the current step. Experiments show that Duet maintains competitive accuracy while achieving 1.34$\times$/2.83$\times$ end-to-end/per-step speedups on LLaDA2.1-mini and 2.41$\times$/4.72$\times$ on SDAR-8B-Chat-b32 at a 32K context length. 
\end{abstract}

\input{sections/1_Introduction}

\input{sections/2_Preliminaries}

\input{sections/3_Method}
\input{sections/4_Experiments}
\input{sections/5_Related_Work}
\input{sections/6_Conclusion}

\section*{Acknowledgments}
This work is supported by
Jiangsu Science and Technology Major Project (\#BG2025005), the Collaborative Innovation Center
of Novel Software Technology and Industrialization, and NanJing Kunpeng\&Ascend Center of Cultivation.
Jingwei Xu (\texttt{jingweix@nju.edu.cn}) is the corresponding author.

\section*{AI use statement}

The core idea and algorithmic design of Duet were developed by the authors.
We used generative AI tools primarily to modify and optimize portions of
the implementation and evaluation code, monitor experimental runs and
inspect logs, and edit the manuscript for clarity and presentation. The
tools also assisted in reviewing the presentation of mathematical claims
and interpreting benchmark results; they did not originate the core method.
The authors reviewed the AI-assisted work and take full responsibility
for the final code, results, and claims in this paper.

\section*{Reproducibility statement}

We release the anonymized source code and testing procedures for Duet at the
repository linked in Section~\ref{sec:experiments}. The release includes the
implementation and evaluation scripts for the short- and long-context
benchmarks and inference-speed measurements. Section~\ref{sec:method}
specifies the inference algorithm, while Section~\ref{sec:experiments}
describes the models, decoding settings, baselines, benchmarks, and evaluation
protocol. The appendix provides task-wise LongBench results together with
detailed latency and decoding-step measurements.
Together, these materials support reproduction of the reported quality and
efficiency comparisons.

\clearpage
\bibliographystyle{plainnat}
\bibliography{references}

\input{sections/7_Appendix}

\end{document}

%% file: math_commands.tex
\usepackage{amsmath,amsfonts,bm}

\def\eqref#1{equation~\ref{#1}}

\def\1{\bm{1}}

\DeclareMathAlphabet{\mathsfit}{\encodingdefault}{\sfdefault}{m}{sl}
\SetMathAlphabet{\mathsfit}{bold}{\encodingdefault}{\sfdefault}{bx}{n}



%% file: sections/1_Introduction.tex
\section{Introduction}

DLMs have recently emerged as an alternative paradigm for text generation, replacing strictly left-to-right autoregressive decoding with iterative denoising over multiple token positions~\citep{sahoo2024simple,nie2025neurips-large,ye2025dream}. By updating several positions in parallel, DLMs offer a promising path toward more flexible and potentially faster generation~\citep{wang2025diffusion}. Recent advances have demonstrated competitive performance across general language understanding, mathematical reasoning, and code generation tasks~\citep{nie2025llada20}, further motivating interest in efficient DLM inference. 

However, the iterative nature of denoising introduces substantial computational redundancy. In block DLM (BDLM) decoding, a current generation block is iteratively refined over multiple denoising steps while conditioning on an generated prefix~\citep{arriola2025iclr-block}. This inference pattern repeatedly incurs two major sources of unnecessary computation. First, every denoising step attends over the full historical prefix, making attention increasingly expensive as the context grows~\citep{long-etal-2026-focus}. Second, all positions in the current block are typically propagated through the full Transformer stack at every denoising step, although only a subset of unresolved positions are ultimately selected for the current update~\citep{liang2026focus}. These two forms of redundancy jointly make BDLM inference increasingly expensive.

Motivated by these inefficiencies, we introduce Duet, a training-free dual-sparse inference framework for BDLMs. Duet addresses the two sources of redundant computation with complementary sparsity mechanisms. First, current-block queries concentrate much of their historical attention on a compact prefix subset that remains stable across denoising steps within the same generation block. Duet exploits this structure through historical \emph{prefix KV sparsity}, selecting a compact historical KV subset once for each generation block and reusing it throughout the block. Second, the positions selected by the final decoding update become predictable from early Transformer representations. Duet leverages this predictability through \emph{early query sparsity} within block, applying the model's original decoding policy at an early selection layer to identify active query rows and propagating only these rows through the remaining layers. Together, the two mechanisms reduce both historical attention and deep token-wise computation, including attention projections, FFN/MoE modules, and other token-wise operations.
\begin{figure}[t]
    \centering
    \includegraphics[width=\linewidth]{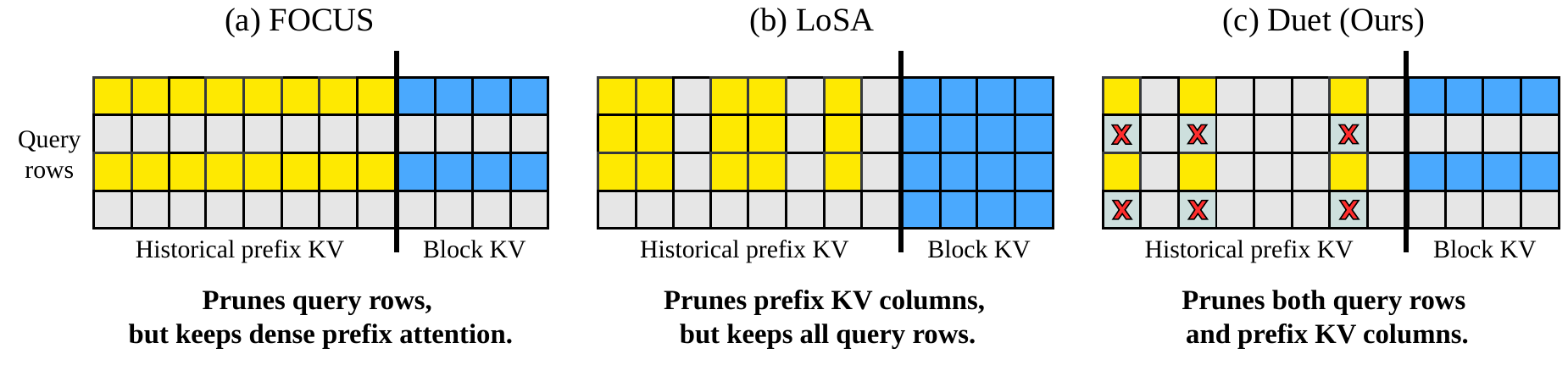}
    \caption{Illustration of different sparsity patterns in BDLM inference. FOCUS only reduces computation along the query dimension, whereas LoSA only sparsifies the historical prefix. Our method is designed around the two orthogonal computation dimensions, selectively reducing both the query rows propagated through deep layers and the historical prefix KV states they attend to.}
    \label{fig:illustration}
\end{figure}
We evaluate Duet on LLaDA2.1-mini~\citep{bie2026llada21speedingtextdiffusion} and SDAR-8B-Chat-b32~\citep{cheng2025sdar} across long-context and standard reasoning benchmarks. Duet consistently reduces inference cost while preserving competitive generation quality. At a 32K context length, Duet achieves 1.34$\times$/2.83$\times$ end-to-end/per-step speedups on LLaDA2.1-mini and 2.41$\times$/4.72$\times$ on SDAR-8B-Chat-b32. We further analyze the two empirical observations underlying Duet, validating that compact historical prefix subsets remain stable within a generation block and that final update positions become highly predictable at early Transformer layers.

Our contributions are summarized as follows:
\begin{itemize}
    \item We identify two complementary sources of redundancy in BDLM inference and show that historical-prefix relevance remains stable within a generation block, while final update positions can be predicted from early Transformer representations.
    \item We propose Duet, a training-free dual-sparse inference framework that combines prefix KV sparsity with early query sparsity to reduce both historical attention and deep token-wise computation.
    \item We conduct extensive experiments on two representative BDLMs across long-context, reasoning, code generation, and knowledge benchmarks, demonstrating competitive generation quality and substantial inference speedups.
\end{itemize}

%% file: sections/2_Preliminaries.tex
\section{Preliminaries}
\label{sec:preliminaries}

\textbf{BDLM Inference.}
DLMs generate text through iterative denoising rather than left-to-right autoregressive decoding. We consider BDLM inference, where generation proceeds one block at a time. Let $P$ denote the length of the currently generated prefix and $B$ the size of the current generation block. At denoising step $s$, the input sequence is
\begin{equation}
\mathbf{x}^{(s)}
=
\left(
\underbrace{x_1,\ldots,x_P \vphantom{x_{P+B}^{(s)}}}_{\text{Prefix}},
\underbrace{x_{P+1}^{(s)},\ldots,x_{P+B}^{(s)}}_{\text{Block}}
\right).
\label{eq:dlm_input}
\end{equation}
After the final layer, the model applies its decoding policy to the current-block representations and resolves a subset of positions, while the unresolved positions are processed again in subsequent denoising steps. Once all positions in the block are resolved, the decoded block is appended to the historical prefix and generation proceeds to the next block.

Within a generation block, the historical prefix remains fixed while the current-block representations evolve across denoising steps. Therefore, historical prefix KV states can be cached and reused, whereas the $B$ current-block positions are repeatedly passed through the Transformer~\citep{wu2025fastdllm,wu2025fastdllmv2}. At layer $l$ and step $s$, we denote the current-block queries by
$\mathbf{Q}^{(l,s)}\in\mathbb{R}^{B\times H_q\times d_h}$ and the
cached historical keys and values by
$\mathbf{K}_{P}^{(l)},\mathbf{V}_{P}^{(l)}
\in\mathbb{R}^{P\times H_{kv}\times d_h}$.
Each current-block query attends to the $P$ cached prefix positions and the $B$ positions in its own block. Suppressing head counts and constant factors, the per-layer, per-step costs of dense attention and the feed-forward (FFN) or mixture-of-experts (MoE) module are
\begin{equation}
T_{\mathrm{attn,dense}}^{(l)}
=O\!\left(B(P+B)d_h+B c_{\mathrm{proj}}^{(l)}\right),
\qquad
T_{\mathrm{ff,dense}}^{(l)}
=O\!\left(B c_{\mathrm{ff}}^{(l)}\right),
\label{eq:dense_layer_cost}
\end{equation}
where $c_{\mathrm{proj}}^{(l)}$ is the per-row attention projection cost and $c_{\mathrm{ff}}^{(l)}$ is the per-row cost of the FFN or activated MoE experts, including routing. The first attention term accounts for query--key interactions and value aggregation. These costs expose two complementary reduction opportunities: attending to fewer historical KV states and propagating fewer current-block query rows through deep Transformer layers.

\textbf{Prefix-Sparse Attention.}
\label{sec:preliminaries_prefix_sparse}
We first consider sparsity along the historical-prefix dimension. For
clarity, consider a single attention head and omit layer and denoising
step superscripts. Let $\mathbf{q}_i\in\mathbb{R}^{d_h}$ be a
current-block query and
$\mathbf{K}_{P},\mathbf{V}_{P}\in\mathbb{R}^{P\times d_h}$ and
$\mathbf{K}_{B},\mathbf{V}_{B}\in\mathbb{R}^{B\times d_h}$ be the
historical-prefix and current-block KV states, respectively. Dense attention computes
\begin{equation}
    \mathbf{o}_i
    =
    \operatorname{softmax}
    \left(
        \frac{
            \mathbf{q}_i[\mathbf{K}_{P};\mathbf{K}_{B}]^{\top}
        }{\sqrt{d_h}}
    \right)
    [\mathbf{V}_{P};\mathbf{V}_{B}].
    \label{eq:dense_prefix_attention}
\end{equation}

Prefix-sparse attention instead retains only a subset
$\mathcal{S}_{P}\subseteq\{1,\ldots,P\}$ of historical positions and
computes
\begin{equation}
    \widetilde{\mathbf{o}}_i
    =
    \operatorname{softmax}
    \left(
        \frac{
            \mathbf{q}_i
            [\mathbf{K}_{P}[\mathcal{S}_{P}];\mathbf{K}_{B}]^{\top}
        }{\sqrt{d_h}}
    \right)
    [\mathbf{V}_{P}[\mathcal{S}_{P}];\mathbf{V}_{B}],
    \label{eq:sparse_prefix_attention}
\end{equation}
where the semicolon concatenates along the KV sequence dimension and
$[\mathcal{S}_{P}]$ gathers the indexed historical rows. The original attention
mask, when present, applies to both expressions. Only the historical KV
portion changes; prefix sparsity does not remove the current-block KV states.
Prefix sparsity therefore targets the context-length-dependent component of
BDLM inference, which becomes increasingly important as $P$ grows.

\textbf{Query-Sparse Deep Computation.}
\label{sec:preliminaries_query_sparse}
BDLM inference also repeatedly propagates all $B$ current-block positions through the Transformer, even though only a subset may require deep computation at a given denoising step. We denote the active block positions at step $s$ by $\mathcal{S}_{Q}^{(s)}\subseteq\{1,\ldots,B\}$ and their number by $Q_s=|\mathcal{S}_{Q}^{(s)}|$. Query sparsity omits the inactive query rows from subsequent layers: these rows do not incur deep attention projections or query-side attention, nor the corresponding FFN/MoE computation. The current block still provides its KV states to the active queries.
\begin{figure}[t]
    \centering
    \includegraphics[width=\linewidth]{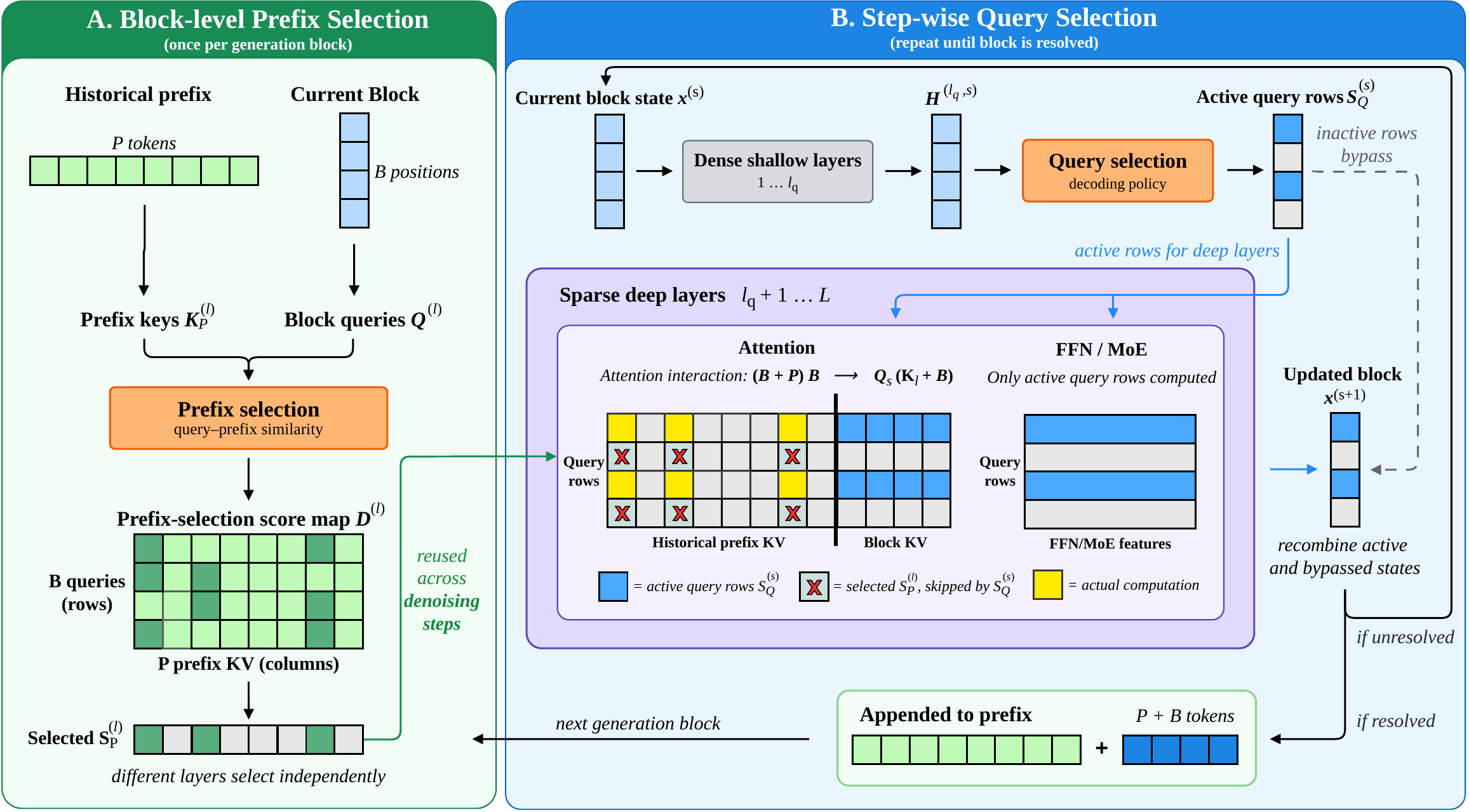}
    \caption{\textbf{Overview of Duet.} \textbf{(A) Block-level prefix selection:} After the first dense step, Duet selects a compact historical prefix subset for each sparse suffix layer and reuses the resulting KV cache within the generation block. \textbf{(B) Step-wise query selection:} in subsequent denoising steps, Duet identifies active query rows at an early selection layer and propagates only these rows through the remaining Transformer layers, reducing redundant computation in both attention and FFN/MoE. Active and bypassed states are then recombined for the next denoising step.}
    \label{fig:overview}
\end{figure}
Thus, prefix sparsity reduces the historical KV context, while query sparsity reduces the number of current-block positions receiving deep computation. The next section describes how Duet combines them during BDLM decoding.

%% file: sections/3_Method.tex
\section{Method}
\label{sec:method}

In this section, we present Duet, a training-free dual-sparse inference framework that reduces redundant computation in BDLM decoding through prefix KV sparsity and early query sparsity. We first provide an overview of Duet, and then describe the two sparsity mechanisms in detail.

\subsection{Overview}
\label{sec:method_overview}

As shown in Figure~\ref{fig:overview}, Duet applies two complementary forms of sparsity to a generation block of $B$ positions conditioned on a prefix of length $P$. The first denoising step ($s=0$) runs densely, providing the block queries and the full historical KV states needed for prefix selection. Duet then constructs a compact prefix cache from these states without an additional dense model forward. In subsequent steps, it processes the block densely through an early query-selection layer $l_Q$ and propagates only active positions through the remaining layers. We use zero-based indices for layers and denoising steps.

For each suffix layer $l>l_Q$, \emph{prefix KV sparsity} uses block queries and historical keys to construct a subset $\mathcal{S}_{P}^{(l)}\subseteq\{1,\ldots,P\}$. This subset remains fixed within the current block; at the next block, selection is repeated over the full historical KV cache. Layers $0,\ldots,l_Q$ retain the full prefix KV states. At each subsequent step $s\geq1$, \emph{early query sparsity} identifies an active set $\mathcal{S}_{Q}^{(s)}\subseteq\{1,\ldots,B\}$ at layer $l_Q$. Only these $Q_s$ rows traverse layers $l_Q+1,\ldots,L-1$, while the other rows bypass the suffix computation.

Within the sparse suffix, active queries attend to the selected historical KV states together with the current block KVs. Restricting the suffix to $\mathcal{S}_{Q}^{(s)}$ also avoids attention projections and FFN/MoE computation for inactive rows. The resulting active states are scattered back to their original block positions and combined with the bypassed states. The process repeats until the block is resolved, after which it is appended to the historical prefix.

\subsection{Prefix KV Sparsity}
\label{sec:method_prefix}

\textbf{Design intuition.}
Within a generation block, the historical prefix remains fixed while the current block is iteratively refined. Empirically, we observe that block queries concentrate a large fraction of their prefix-attention mass on a compact subset of historical tokens, and that this subset remains largely stable throughout the denoising steps within the same block. This motivates selecting a compact historical context once at the beginning of each generation block and reusing the resulting KV cache throughout that block. We validate this block-level stability in Section~\ref{sec:analysis_prefix}.

\textbf{Query--prefix similarity.}
Prefix selection is performed for each suffix layer $l>l_Q$, using queries captured during the first dense step. Let $\mathbf{q}_{h,i}^{(l,0)}\in\mathbb{R}^{d_h}$ be the attention-space query at block position $i$ and query head $h$, and let $\mathbf{k}_{g(h),j}^{(l)}\in\mathbb{R}^{d_h}$ be the historical key at position $j$ in its corresponding KV head $g(h)$ under grouped-query attention. We measure their similarity by
\begin{equation}
    a_{hij}^{(l)}
    =
    \left\langle
        \mathbf{q}_{h,i}^{(l,0)},
        \mathbf{k}_{g(h),j}^{(l)}
    \right\rangle .
    \label{eq:prefix_similarity}
\end{equation}
\textbf{Block-shared prefix selection.} Rather than constructing a separate historical context for each query, Duet selects one prefix subset shared across the block and query heads. Given a target budget $B_P$, each of the $BH_q$ query-head rows selects its $m=\max(1,\lceil B_P/(BH_q)\rceil)$ highest-scoring historical positions, and the candidates are merged. We score each historical position across all block queries and query heads by
\begin{equation}
    r_j^{(l)}
    =
    \max_{\substack{1\leq i\leq B\\1\leq h\leq H_q}}
    a_{hij}^{(l)},
    \label{eq:prefix_aggregate_score}
\end{equation}
which captures the strongest match to any query-head row. If the candidate union contains fewer than $\min(B_P,P)$ distinct positions, we fill it with the highest-scoring remaining positions according to $r_j^{(l)}$; if it contains more, we retain the $\min(B_P,P)$ highest-scoring candidates. Thus, under a strict prefix budget, the shared set has size $K_l=|\mathcal{S}_P^{(l)}|=\min(B_P,P)$. Ranking by the unscaled dot product is equivalent to ranking by the scaled attention logit for a fixed head dimension. Scoring the full prefix costs $O(BP d_h)$ per suffix layer, suppressing head counts, and is performed once per generation block.

The selected indices gather the cached keys and values, $\widetilde{\mathbf{K}}_{P}^{(l)}=\mathbf{K}_{P}^{(l)}[\mathcal{S}_{P}^{(l)}]$ and $\widetilde{\mathbf{V}}_{P}^{(l)}=\mathbf{V}_{P}^{(l)}[\mathcal{S}_{P}^{(l)}]$. This compact cache is reused throughout the current block; the full historical cache remains available for selection at the next block.

\subsection{Early Query Sparsity}
\label{sec:method_query}

\textbf{Design intuition.} At each denoising step, standard BDLM inference propagates all positions in the current generation block through every Transformer layer, even though only a subset of unresolved positions are ultimately selected for the current update. We observe that these update positions can already be predicted from early Transformer representations. This suggests that query selection can be performed before the final layer, allowing Duet to avoid redundant deep computation on positions unlikely to be updated. Accordingly, Duet designates an early Transformer layer $l_Q$ as the selection layer and propagates only the predicted active query rows through the remaining deep layers. We empirically validate this early predictability in Section~\ref{sec:analysis_query}.

\textbf{Early prediction of query positions.} At a subsequent denoising step $s\geq1$, Duet evaluates layers $0,\ldots,l_Q$ densely over the current block, including attention to the full historical prefix, and obtains the intermediate representations $\mathbf{H}^{(l_Q,s)}$. The model's original decoding rule is then applied to $\mathbf{H}^{(l_Q,s)}$ to estimate which masked positions are likely to be selected for the current update. Let $\mathcal{U}^{(s)}$ denote the unresolved positions at step $s$. The query retention ratio $\rho_Q$ sets a nominal target of
$\lceil \rho_Q |\mathcal{U}^{(s)}| \rceil$ active unresolved queries,
selected according to the model's decoding policy (confidence-based for
LLaDA and in block order for sequential SDAR); previously resolved
positions are always retained as well. Together, these positions form $\mathcal{S}_Q^{(s)}$, whose $Q_s$ rows are propagated through the suffix layers. In this way, the retention ratio controls the amount of deep computation devoted to unresolved positions, while reusing the original decoding rule keeps early query selection consistent with the underlying BDLM inference procedure without introducing an additional learned predictor.



\textbf{Sparse suffix computation.} For $s\geq1$ and $l>l_Q$, only the $Q_s$ active query rows attend to the compact historical KV cache indexed by $\mathcal{S}_P^{(l)}$ and the $B$ current-block KV positions. The first dense step initializes the current-block KV cache at all $B$
positions. Subsequent sparse steps update the selected entries and reuse
the others, so active queries still attend to all $B$ block positions. Using the cost conventions of Equation~\ref{eq:dense_layer_cost}, the per-layer sparse costs are
\begin{equation}
T_{\mathrm{attn,sparse}}^{(l,s)}
=O\!\left(Q_s(K_l+B)d_h+Q_s c_{\mathrm{proj}}^{(l)}\right),
\qquad
T_{\mathrm{ff,sparse}}^{(l,s)}
=O\!\left(Q_s c_{\mathrm{ff}}^{(l)}\right).
\label{eq:sparse_layer_cost}
\end{equation}
Compared with dense computation, the suffix attends over $K_l+B$ rather than $P+B$ KV positions and processes $Q_s$ rather than $B$ rows through attention projections and FFN/MoE. These expressions apply only after the first dense step and beyond $l_Q$; the first step and the preceding layers retain the dense costs of Equation~\ref{eq:dense_layer_cost}. Because selection at the next block revisits the full prefix, the shorter attended KV sequence does not imply a reduction in the stored full historical KV cache.

\textbf{Position-preserving update.} Sparse suffix computation preserves the
original block indices of all active query rows, allowing their updated
representations to be written back to the corresponding positions after the
deep layers. At the final decoding stage, the original decoding rule is
applied only to the active unresolved positions
$\mathcal{S}_Q^{(s)}\cap\mathcal{U}^{(s)}$. Unselected unresolved positions
bypass the suffix computation, remain masked at this step, and are
reconsidered at the next step. Thus, the per-step update set may differ from
dense decoding. The updated active rows, bypassed unresolved rows, and
previously resolved rows are recombined to reconstruct the full block state
for the next denoising step. Query selection is repeated until all positions
are resolved. Once the current block is fully decoded, it is appended to the
historical context, and Duet proceeds to the next generation block with an
updated prefix.

%% file: sections/4_Experiments.tex
\section{Experiments}
\label{sec:experiments}

In this section, we evaluate Duet in terms of generation quality and inference efficiency across multiple models and benchmarks. We then analyze the two empirical observations underlying its design: the stability of prefix sparsity and the early predictability of query positions.

\subsection{Experimental Setup}
\label{sec:exp_setup}
\begin{figure}[t]
    \centering
    \includegraphics[width=0.9\linewidth]{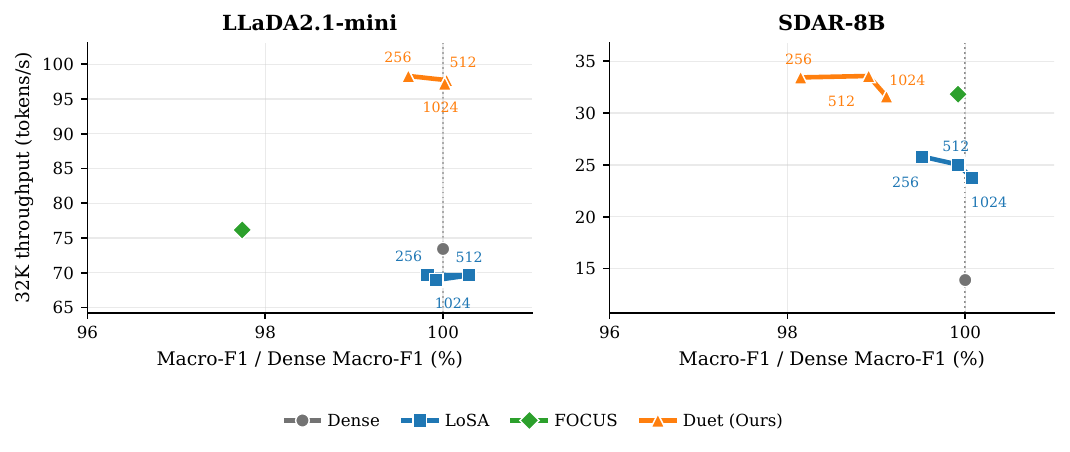}
    \caption{\textbf{LongBench quality--throughput trade-off.}
    Five-task Macro-F1 relative to each model's dense score (100\%)
    versus 32K end-to-end throughput. Absolute Macro-F1 scores are
    reported in Table~\ref{tab:longbench_summary}; task-wise results
    appear in Appendix Table~\ref{tab:longbench_detail}.}
    \label{fig:longbench_tradeoff}
\end{figure}

\begin{table}[t]
    \centering
    \caption{\textbf{LongBench Macro-F1 (\%).}
    Column headings give nominal prefix budgets. For LoSA,
    parentheses give measured prefix KV counts, which can exceed
    those budgets. Higher is better.}
    \label{tab:longbench_summary}
    {\footnotesize
        \setlength{\tabcolsep}{2.5pt}%
        \resizebox{\linewidth}{!}{%
        \begin{tabular}{@{}l*{8}{c}@{}}
        \toprule
        & & & \multicolumn{3}{c}{LoSA}
            & \multicolumn{3}{c}{Duet (Ours)} \\
        \cmidrule(lr){4-6}\cmidrule(l){7-9}
        Model & Dense & FOCUS & 256 & 512 & 1024 & 256 & 512 & 1024 \\
        \midrule
        LLaDA2.1-mini
        & 51.33 & 50.17
        & 51.24 (434) & 51.48 (833) & 51.29 (1566)
        & 51.13 & 51.35 & 51.34 \\
        SDAR-8B
        & 49.69 & 49.65
        & 49.45 (405) & 49.65 (794) & 49.73 (1507)
        & 48.77 & 49.15 & 49.25 \\
        \bottomrule
        \end{tabular}%
        }\par}
\end{table}

\begin{table}[t]
    \centering
    \small
    \caption{Generation quality on short-context benchmarks. We report accuracy (\%) on GSM8K, MMLU, and MATH-500, and pass@1 (\%) on HumanEval. For LoSA, parentheses give the average number of selected KV tokens per task. Higher is better.}
    \label{tab:shortbench_quality}
    \begin{tabular}{llccccc}
        \toprule
        Model & Method & Budget & GSM8K & MMLU & HumanEval & MATH-500 \\
        \midrule
        \multirow{4}{*}{LLaDA2.1-mini}
        & Dense & --  & \underline{85.82} & \textbf{67.79} & 78.66 & \underline{91.80} \\
        & LoSA  & 256 & 84.61 (216) & \underline{66.99} (873) & 70.73 (417) & 88.80 (512) \\
        & FOCUS & --  & \textbf{86.05} & 66.76 & \underline{79.88} & \textbf{92.60} \\
        & Duet (Ours)  & 256 & 84.31 & 66.90 & \textbf{80.49} & 87.60 \\
        \midrule
        \multirow{4}{*}{SDAR-8B}
        & Dense & --  & \textbf{88.25} & 74.52 & 61.59 & \underline{63.60} \\
        & LoSA  & 256 & 87.11 (188) & \textbf{74.55} (458) & 54.88 (332) & 58.40 (294) \\
        & FOCUS & --  & 86.88 & 74.52 & \textbf{64.63} & \textbf{65.80} \\
        & Duet (Ours)  & 256 & \underline{87.64} & \underline{74.53} & \underline{62.20} & \underline{63.60} \\
        \bottomrule
    \end{tabular}
\end{table}

\begin{figure}[t]
    \centering
    \includegraphics[width=\linewidth]{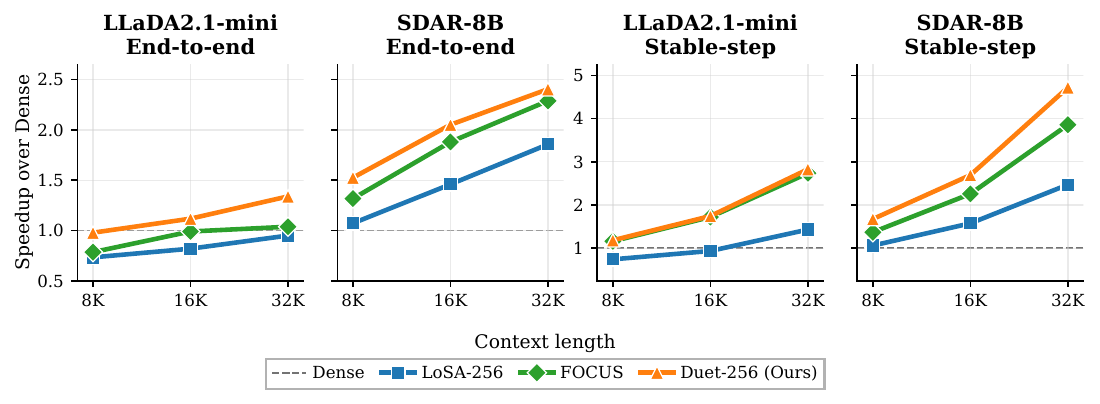}
    \caption{\textbf{Inference speedup across context lengths.}
    End-to-end and stable-step speedups over dense inference on LLaDA2.1-mini and SDAR-8B at 8K, 16K, and 32K context lengths.
    Stable-step latency averages all denoising steps except the first step, highlighting the execution efficiency of sparse decoding.
    Duet achieves increasingly larger speedups as the context grows and consistently provides the strongest stable-step acceleration among the compared methods.
    }
    \label{fig:context_scaling_speedup}
\end{figure}

\textbf{Models.} We evaluate Duet on two representative BDLMs, LLaDA2.1-mini~\citep{bie2026llada21speedingtextdiffusion} and SDAR-8B-Chat-b32~\citep{cheng2025sdar}. Because the predictability of decoding positions emerges at different depths across models, we use model-specific query-sparsity configurations. For LLaDA2.1-mini, we set the query retention ratio to $\rho_Q=0.7$ and the selection layer to $l_Q=1$, while retaining its original decoding policy. For SDAR-8B-Chat-b32, we set $\rho_Q=0.5$ and $l_Q=5$, and use sequential decoding with 16 denoising steps. At the selection layer, Duet applies the same decoding policy used for the final update to predict active query rows.

\textbf{Benchmarks.} We evaluate accuracy on both long- and shorter-context tasks. For long-context evaluation, we use LongBench~\citep{bai2024longbench}, which spans a diverse set of long-document understanding and generation tasks. We additionally evaluate mathematical reasoning on GSM8K~\citep{cobbe2021gsm8k} and MATH-500~\citep{hendrycksmath2021}, code generation on HumanEval~\citep{chen2021evaluating}, and knowledge-intensive question answering on MMLU~\citep{hendryckstest2021}.

\textbf{Baselines and implementation.} We compare Duet against dense inference and two sparse-inference baselines, FOCUS~\citep{liang2026focus} and LoSA~\citep{xi2026losa}. FOCUS primarily sparsifies computation within the current generation block, whereas LoSA focuses on sparsifying attention over the historical prefix. Both baselines use their default configurations. To ensure a fair systems comparison, we implement all methods within the same inference framework and preserve the algorithmic behavior described in the original papers. We further optimize the core sparse operators with Triton kernels and use the same runtime stack across methods. Since the original FOCUS implementation is optimized for shorter-context, larger-batch settings, we adapt its execution path to long-context decoding while keeping its selection and sparse-computation logic unchanged.

\textbf{Efficiency evaluation.} We evaluate inference efficiency on 80 NarrativeQA samples at 8K, 16K, and 32K context lengths with a fixed generation length of 256 tokens on an NVIDIA H800 GPU. We report end-to-end and stable per-step speedups in the main text, with detailed results provided in the appendix; stable per-step latency excludes the first denoising step to remove initialization overhead.

\subsection{Main Results}

\textbf{Generation quality.}
Figure~\ref{fig:longbench_tradeoff} shows the LongBench quality--throughput trade-off, Table~\ref{tab:longbench_summary} reports the corresponding Macro-F1 scores, and Table~\ref{tab:shortbench_quality} reports short-context quality. At nominal budget 512, Duet matches dense Macro-F1 on LLaDA2.1-mini (51.35 vs.\ 51.33) with a $1.33\times$ end-to-end speedup. On SDAR-8B, it reaches 49.15 versus 49.69 with a $2.42\times$ speedup. LoSA also preserves quality, but at nominal budget 256 its measured prefix KV use averages 434 and 405 tokens on the two models, respectively, exceeding the configured budget. Full task-wise scores and KV counts are in Appendix Table~\ref{tab:longbench_detail}.

Duet also maintains comparable quality on shorter-context benchmarks. Across GSM8K, MMLU, HumanEval, and MATH-500, both LLaDA2.1-mini and SDAR-8B closely track their dense counterparts, with LLaDA2.1-mini additionally outperforming Dense on HumanEval. These results show that Duet preserves generation quality across diverse tasks while substantially reducing inference cost, as evaluated next.

\textbf{Inference efficiency.}
As shown in Figure~\ref{fig:context_scaling_speedup} and Table~\ref{tab:efficiency_detail}, Duet provides increasingly larger end-to-end gains as the context grows. On LLaDA2.1-mini, Duet with a prefix budget of 256 achieves a $1.12\times$ speedup at 16K and a $1.34\times$ speedup at 32K, while FOCUS provides only a $1.04\times$ speedup and LoSA remains slower than dense inference at 32K. The same trend is more pronounced on SDAR-8B: Duet reaches $1.53\times$, $2.05\times$, and $2.41\times$ end-to-end speedups at 8K, 16K, and 32K, respectively. These results show that Duet remains effective as the historical prefix grows, where repeatedly attending to dense historical KV states becomes increasingly expensive.

The stable-step results in Table~\ref{tab:efficiency_detail} further isolate the computation saved during sparse decoding. At 32K, Duet reduces the stable-step latency from 58.12\,ms to 20.55\,ms on LLaDA2.1-mini, corresponding to a $2.83\times$ speedup, and from 99.53\,ms to 21.09\,ms on SDAR-8B, corresponding to a $4.72\times$ speedup. In comparison, LoSA reaches only $1.43\times$ and $2.47\times$ speedups on the two models at 32K, respectively, indicating that its prefix-selection overhead substantially limits the benefit of sparse attention. FOCUS achieves stable-step speedups closer to Duet---$2.74\times$ on LLaDA2.1-mini and $3.86\times$ on SDAR-8B at 32K---but this advantage translates less effectively to end-to-end generation. FOCUS changes the decoding trajectory and requires more denoising steps to generate the same number of output tokens; we quantify this effect in Table~\ref{tab:efficiency_detail}. As a result, its end-to-end speedup is lower than what its per-step latency alone would suggest.

\subsection{Analysis}
We analyze stability of prefix sparsity and early predictability of query positions on LLaDA2.1-mini.
\begin{figure}[t]
    \centering

    \begin{subfigure}[t]{0.48\linewidth}
        \centering
        \includegraphics[width=\linewidth]{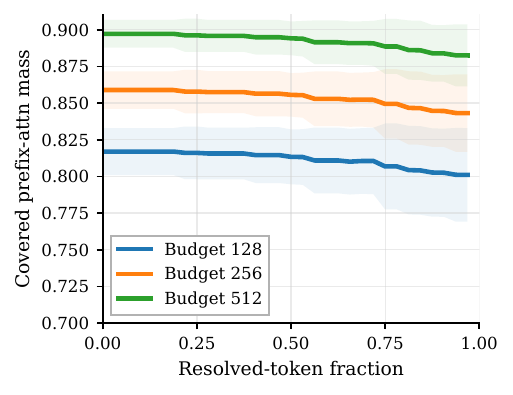}
        \caption{Persistent prefix-attention concentration.}
        \label{fig:prefix_mass}
    \end{subfigure}
    \hfill
    \begin{subfigure}[t]{0.48\linewidth}
        \centering
        \includegraphics[width=\linewidth]{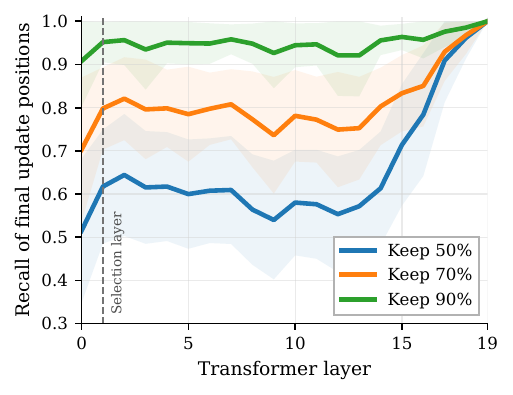}
        \caption{Early predictability of query positions.}
        \label{fig:query_recall}
    \end{subfigure}

    \caption{
        \textbf{Two sparsity opportunities in BDLM inference.}
        (a) A compact prefix subset selected at the beginning of a
        generation block continues to cover most of the prefix-attention
        mass throughout denoising.
        (b) Intermediate representations can identify, with high recall,
        the positions selected by the final-layer decoding policy.
        Results are shown on LLaDA2.1-mini as a representative example.
    }
    \label{fig:motivation}
\end{figure}

\textbf{Stability of prefix sparsity.}
\label{sec:analysis_prefix}
We examine whether a compact prefix subset selected at the beginning of a generation block remains representative throughout the subsequent denoising process. Take LLaDA2.1-mini as the For each block, we keep the selected historical prefix subset fixed and measure the fraction of prefix-attention mass covered by this subset as the block is progressively resolved. Figure~\ref{fig:prefix_mass} shows that even a prefix budget of 256 consistently captures around 85\% of the historical attention mass, while increasing the budget to 512 raises the coverage to nearly 90\%. More importantly, the covered attention mass remains stable throughout block decoding: across all evaluated budgets, the variation over the denoising trajectory is within approximately 2.5 percentage points. These results indicate that the current-block queries maintain a largely stable focus on a compact portion of the historical prefix, supporting our design of selecting prefix KV states once before decoding each block and reusing the resulting cache throughout its denoising steps.

\textbf{Early predictability of query positions.}
\label{sec:analysis_query}
We analyze whether the positions selected by the final decoding update can already be predicted from early Transformer representations. At each denoising step, we apply the model's original decoding policy to hidden states from different layers and measure the recall of the positions selected by the final-layer decoding result. Figure~\ref{fig:query_recall} shows that the final update positions become highly predictable at surprisingly early layers. At layer 1, retaining 70\% of the masked positions already recovers approximately 80\% of the final update positions, while a 90\% retention ratio reaches about 95\% recall. The recall remains high through most of the network and approaches 100\% near the final layers. These results demonstrate that the locations of the final decoding updates can be identified well before the full Transformer depth is evaluated, providing direct support for selecting active query rows at an early layer and avoiding unnecessary deep computation on the remaining positions.


%% file: sections/5_Related_Work.tex
\section{Related Work}

\textbf{DLMs.}
DLMs generate text by iteratively denoising discrete token sequences. Prior work explored discrete-state diffusion, continuous diffusion in embedding space, and diffusion with pretrained masked language models~\citep{austin2021structured,li2022diffusionlm,he2023diffusionbert}. Masked discrete diffusion provided a scalable formulation, later demonstrated at scale by LLaDA and Dream~\citep{sahoo2024simple,nie2025neurips-large,ye2025dream}. Fully parallel masked DLMs predict multiple positions per step, enabling bidirectional context modeling and parallel generation. However, repeatedly processing the full sequence incurs redundant computation on largely unchanged tokens, while bidirectional attention prevents direct reuse of the append-only KV cache used by autoregressive models~\citep{wu2025fastdllm,liu2025dllmcache}.

\textbf{From DLMs to BDLMs.}
Fast-dLLM demonstrates the potential of block-level organization for DLM acceleration through block-wise approximate KV caching and confidence-aware parallel decoding~\citep{wu2025fastdllm}. Subsequent methods explore adaptive cache reuse at different granularities, including segment-level updates, delayed updates, and token-level state selection~\citep{liu2025dllmcache,ma2025dkvcache,jiang2025d2cache}. BDLMs make this structure part of the generation paradigm: Block Diffusion generates blocks autoregressively while refining tokens through diffusion within each block~\citep{arriola2025iclr-block}; SDAR adapts pretrained autoregressive models to block-wise diffusion~\citep{cheng2025sdar}; and LLaDA2.0 and LLaDA2.1 extend the scale and token-editing capabilities of this model family~\citep{nie2025llada20,bie2026llada21speedingtextdiffusion}. Fast-dLLM v2 further combines block-diffusion model design with hierarchical caching and parallel decoding~\citep{wu2025fastdllmv2}.

\textbf{Inference optimization for existing BDLMs.}
Recent work optimizes existing BDLMs, including SDAR and LLaDA2.x, by reducing current-block computation or historical-prefix attention~\citep{liang2026focus,xi2026losa}. FOCUS estimates token-wise decodability to skip positions unlikely to be updated, but under confidence-based decoding this may increase the number of denoising steps, limiting end-to-end gains~\citep{liang2026focus}. LoSA sparsifies long-context prefix attention by recomputing attention for active queries and reusing outputs for stable ones, while repeated query-dependent KV selection still incurs nontrivial overhead~\citep{xi2026losa}. As a result, the reduction in attention computation does not always translate into substantial end-to-end speedup.

%% file: sections/6_Conclusion.tex
\section{Conclusion}

We presented Duet, a training-free dual-sparse framework for BDLM inference that combines block-wise prefix KV selection with early query selection to avoid repeated historical attention and deep processing of inactive positions. The selected prefix is reused across denoising steps, while only predicted active queries traverse the remaining Transformer layers. Duet maintains competitive accuracy across long-context and standard benchmarks, achieving 1.34$\times$/2.83$\times$ end-to-end/per-step speedups on LLaDA2.1-mini and 2.41$\times$/4.72$\times$ on SDAR-8B-Chat-b32 at 32K. Stable prefix relevance within blocks and early predictability of update positions support both sparsity mechanisms.z



%% file: sections/7_Appendix.tex
\clearpage

\appendix
\section{Efficiency Evaluations}

We evaluate inference efficiency on 80 NarrativeQA samples at 8K, 16K, and
32K context lengths, generating 256 tokens per sample on an NVIDIA H800 GPU.
Table~\ref{tab:efficiency_detail} reports three complementary measurements:
mean end-to-end (E2E) generation latency, stable-step latency excluding the
first denoising step, and the mean number of decoding steps per sample.
End-to-end latency includes prefill and generation overheads, so its speedup
need not equal the stable-step speedup.

\begin{table}[ht]
\centering
\caption{\textbf{Detailed inference efficiency results.}
E2E is end-to-end latency in seconds; Step is stable-step latency in
milliseconds; Steps is the mean number of decoding steps per sample.
Lower latency is better.}
\label{tab:efficiency_detail}
\begingroup
\scriptsize
\setlength{\tabcolsep}{2.6pt}
\renewcommand{\arraystretch}{1.12}
\resizebox{\linewidth}{!}{%
\begin{tabular}{@{}ll*{9}{c}@{}}
\toprule
& & \multicolumn{3}{c}{8K context}
    & \multicolumn{3}{c}{16K context}
    & \multicolumn{3}{c}{32K context} \\
\cmidrule(lr){3-5}\cmidrule(lr){6-8}\cmidrule(l){9-11}
Method & Budget
& E2E (s) & Step (ms) & Steps
& E2E (s) & Step (ms) & Steps
& E2E (s) & Step (ms) & Steps \\
\midrule
\multicolumn{11}{@{}l}{\textit{LLaDA2.1-mini}} \\
\midrule
Dense & -- & 0.841 & 24.03 & 16.06 & 1.447 & 35.47 & 16.00 & 3.486 & 58.12 & 25.05 \\
FOCUS & -- & 1.072 & 20.65 & 24.05 & 1.462 & 20.58 & 25.15 & 3.361 & 21.23 & 64.53 \\
LoSA & 256 & 1.148 & 32.25 & 16.65 & 1.767 & 37.78 & 16.18 & 3.672 & 40.54 & 27.81 \\
     & 512 & 1.171 & 33.12 & 16.59 & 1.774 & 38.00 & 16.18 & 3.676 & 40.89 & 28.06 \\
     & 1024 & 1.188 & 34.25 & 16.51 & 1.789 & 39.09 & 16.10 & 3.712 & 40.71 & 28.81 \\
\textbf{Duet(Ours)} & 256 & 0.860 & 20.26 & 15.60 & 1.295 & 20.32 & 15.88 & 2.604 & 20.55 & 23.29 \\
     & 512 & 0.840 & 19.54 & 15.58 & 1.292 & 20.22 & 15.94 & 2.620 & 20.57 & 24.08 \\
     & 1024 & 0.871 & 20.24 & 15.48 & 1.303 & 20.36 & 15.70 & 2.633 & 20.46 & 24.16 \\
\midrule
\multicolumn{11}{@{}l}{\textit{SDAR-8B}} \\
\midrule
Dense & -- & 5.350 & 35.42 & 128 & 9.279 & 56.88 & 128 & 18.416 & 99.53 & 128 \\
FOCUS & -- & 4.062 & 25.80 & 128 & 4.934 & 25.18 & 128 & 8.046 & 25.80 & 128 \\
LoSA & 256 & 4.992 & 33.30 & 128 & 6.360 & 36.05 & 128 & 9.920 & 40.26 & 128 \\
     & 512 & 5.252 & 35.37 & 128 & 6.616 & 38.33 & 128 & 10.227 & 42.46 & 128 \\
     & 1024 & 5.684 & 38.76 & 128 & 7.050 & 41.77 & 128 & 10.798 & 46.64 & 128 \\
\textbf{Duet(Ours)} & 256 & 3.508 & 21.12 & 128 & 4.523 & 21.12 & 128 & 7.656 & 21.09 & 128 \\
     & 512 & 3.524 & 21.21 & 128 & 4.563 & 21.46 & 128 & 7.624 & 20.84 & 128 \\
     & 1024 & 3.526 & 21.05 & 128 & 4.588 & 20.96 & 128 & 8.100 & 20.58 & 128 \\
\bottomrule
\end{tabular}%
}
\endgroup
\end{table}

At 32K context length, Duet with a prefix budget of 256 reduces
LLaDA2.1-mini's stable-step latency from 58.12 to 20.55\,ms and its
end-to-end latency from 3.486 to 2.604\,s. Its mean step count decreases
from 25.05 to 23.29. FOCUS reaches a similar stable-step latency of
21.23\,ms but takes 64.53 steps on average, partly explaining its smaller
end-to-end gain. SDAR-8B uses its default sequential decoding policy, which
fixes the number of decoding steps for a given generation length. Thus, all
methods take 128 steps in this experiment. On SDAR-8B, Duet reduces
stable-step latency from 99.53 to 21.09\,ms and end-to-end latency from
18.416 to 7.656\,s at 32K.

\clearpage
\section{Detailed LongBench Results}

Table~\ref{tab:longbench_detail} reports the full task-wise F1 results
underlying the LongBench Macro-F1 values in
Figure~\ref{fig:longbench_tradeoff}. Each task contributes equally to
Macro-F1. LoSA's parenthesized values are measured averages of attended
prefix KV tokens, not its configured budget; the parenthesized value in
the Macro-F1 column is the mean of the five task-level KV counts.

\begin{table}[ht]
    \centering
    \caption{Task-wise results on LongBench. We report F1 (\%) on five long-context tasks and Macro-F1, the unweighted average across tasks. For LoSA, parentheses give the measured average number of attended prefix KV tokens per task; the Macro-F1 column gives the mean of these five KV counts. Higher F1 is better.}
    \label{tab:longbench_detail}
    \resizebox{\linewidth}{!}{
    \begin{tabular}{llcccccc}
        \toprule
        Method & Budget &
        HotpotQA & TriviaQA & NarrativeQA & Qasper & MultiFieldQA & Macro-F1 \\
        \midrule
        \multicolumn{8}{l}{\textit{LLaDA2.1-mini}} \\
        \midrule
        Dense & --   & 54.60 & 85.91 & 24.71 & 42.15 & 49.26 & 51.33 \\
        FOCUS & --   & 55.23 & 84.74 & 21.96 & 41.89 & 47.03 & 50.17 \\
        LoSA & 256 & 54.74 (430) & 85.51 (426) & 23.87 (447) & 43.20 (435) & 48.90 (433) & 51.24 (434) \\
             & 512 & 56.13 (833) & 85.13 (820) & 23.23 (873) & 43.32 (813) & 49.60 (825) & 51.48 (833) \\
             & 1024 & 55.60 (1594) & 85.86 (1549) & 23.02 (1677) & 42.80 (1483) & 49.15 (1529) & 51.29 (1566) \\
        Duet (Ours) & 256 & 54.02 & 84.40 & 24.69 & 43.27 & 49.29 & 51.13 \\
                    & 512 & 55.36 & 85.11 & 24.98 & 42.29 & 49.01 & 51.35 \\
                    & 1024 & 55.13 & 85.45 & 23.97 & 42.90 & 49.25 & 51.34 \\
        \midrule
        \multicolumn{8}{l}{\textit{SDAR-8B}} \\
        \midrule
        Dense & --   & 52.11 & 83.15 & 25.03 & 34.85 & 53.29 & 49.69 \\
        FOCUS & --   & 51.55 & 83.65 & 24.78 & 34.93 & 53.34 & 49.65 \\
        LoSA & 256 & 51.96 (412) & 83.12 (405) & 25.25 (398) & 34.38 (403) & 52.56 (406) & 49.45 (405) \\
             & 512 & 51.92 (809) & 83.12 (784) & 25.31 (804) & 34.49 (785) & 53.42 (786) & 49.65 (794) \\
             & 1024 & 51.93 (1546) & 83.20 (1490) & 25.37 (1571) & 34.47 (1461) & 53.66 (1467) & 49.73 (1507) \\
        Duet (Ours) & 256 & 50.34 & 83.37 & 24.91 & 34.37 & 50.85 & 48.77 \\
                    & 512 & 51.11 & 83.12 & 24.76 & 34.74 & 52.01 & 49.15 \\
                    & 1024 & 50.87 & 82.87 & 25.87 & 34.78 & 51.86 & 49.25 \\
        \bottomrule
    \end{tabular}
    }
\end{table}